\documentclass[unnumsec,webpdf,contemporary,large,refnumbers]{oup-authoring-template}%

\graphicspath{{Fig/}}

\theoremstyle{thmstyleone}%
\theoremstyle{thmstyletwo}%
\theoremstyle{thmstylethree}%

\begin{document}

\journaltitle{PNAS Nexus (preprint)}
\DOI{\url{https://doi.org/10.1093/pnasnexus/pgac039}}
\copyrightyear{2022}
\pubyear{2022}
\access{}
\appnotes{Preprint}

\firstpage{1}


\title[Ranking the information content of distance measures]{Ranking the information content of distance measures}

\author[a,b]{Aldo Glielmo}
\author[a]{Claudio Zeni} 
\author[c]{Bingqing Cheng}
\author[d]{G\'abor Cs\'anyi}
\author[a,$\ast$]{Alessandro Laio}

\authormark{Glielmo et al.}

\address[a]{\orgdiv{Physics Department}, \orgname{International School for Advanced Studies (SISSA)}, \orgaddress{\street{Via Bonomea 265}, \postcode{34136}, \state{Trieste}, \country{Italy}}}

\address[b]{\orgname{Banca d'Italia}, \orgaddress{\country{Italy}}\textsuperscript{\dag} }

\address[c]{\orgname{The Institute of Science and Technology Austria}, \orgaddress{\street{Am Campus 1}, \postcode{3400}, \state{Klosterneuburg}, \country{Austria} }}

\address[d]{\orgdiv{Engineering Laboratory}, \orgname{University of Cambridge}, \orgaddress{\street{Trumpington St}, \postcode{CB21PZ}, \state{Cambridge}, \country{United Kingdom}}}

\corresp[$\ast$]{To whom correspondence should be addressed: \href{email:laio@sissa.it}{laio@sissa.it} \, }

\corresp[\dag]{The views and opinions expressed in this paper are those of the authors and do not necessarily reflect the official policy or
position of Banca d’Italia.}


\editor{Associate Editor: Name}

\abstract{
Real-world data typically contain a large number of features that are often heterogeneous in nature, relevance, and also units of measure.
When assessing the similarity between data points, one can build various distance measures using subsets of these features.
Finding a small set of features that still retains sufficient information about the dataset is important for the successful application of many statistical learning approaches.
We introduce a statistical test that can assess the relative information retained when using two different distance measures, and determine if they are equivalent, independent, or if one is more informative than the other.
This ranking can in turn be used to identify the most informative distance measure and, therefore, the most informative set of features, out of a pool of candidates.
To illustrate the general applicability of our approach, we show that it reproduces the known importance ranking of policy variables for Covid-19 control, and also identifies compact yet informative descriptors for atomic structures.
We further provide initial evidence that the information asymmetry measured by the proposed test can be used to infer relationships of causality between the features of a dataset.
The method is general and should be applicable to many branches of science.
}

\boxedtext{
%
In real-world data sets many characteristics are often associated with each data point, and one can imagine different ways to define the similarity between two samples.
For example, in a clinical database two patients might be compared based on their age, sex, or height, or on the results of specific clinical exams.
In this work we introduce a method which allows studying the relationship between different distances (or similarity) measures defined on the same dataset.
One can find that two distances are unrelated, that they bring equal information, or that one of the two distances allows predicting the other, while the reverse is not true.
This allows finding distances which are maximally informative for a prediction, and detecting causality relationships.}

\maketitle
\section{Introduction}

An open challenge in machine learning is to extract useful information from a database with relatively few data points, but with a large number of features available for each point~\cite{Ni2020_Few_shot_learning_survey,Lopes2017_Facial_expression_few_data,Valera2020_Handling_heterogeneous_data_with_VAEs}.
For example, clinical databases typically include data for a few hundreds patients with a similar clinical history, but an enormous amount of information for each patient: the results of clinical exams, imaging data, and a record of part of their genome \cite{Biobank2015}.
In cheminformatics and materials science, molecules and materials can be described by a large number of features, but databases are limited in size by the great cost of the  calculations or the experiments required to predict quantum properties~\cite{Pande2017_Low_data_drug_discovery,Yoshida2019_Predicting_materials_with_little_data}.
In short, real-world data are often ``big data'', but in the wrong direction: instead of millions of data points, there are often too many features for a few samples.
As such, training accurate learning models is challenging, and even more so when using deep neural networks, which typically require a large amount of independent samples \cite{Shorten2019_Survey_data_augmentation}.

One way to circumvent this problem is to perform preliminary feature selection, and discard features that appear irrelevant or redundant \cite{Yang2018_Feature_selection_new_perspective,Lopes2017_Facial_expression_few_data,Bogunovic_Feature_selection_review,Deng_Feature_selection_review}.
Alternatively, one can perform a dimensional reduction aimed at finding a representation of the data with few variables built as functions of the original features \cite{vanderMaaten:2008tm, mcinnes2018umap, Bengio:2013bu}.

\begin{figure*}
\centering
\includegraphics[width=0.9\textwidth]{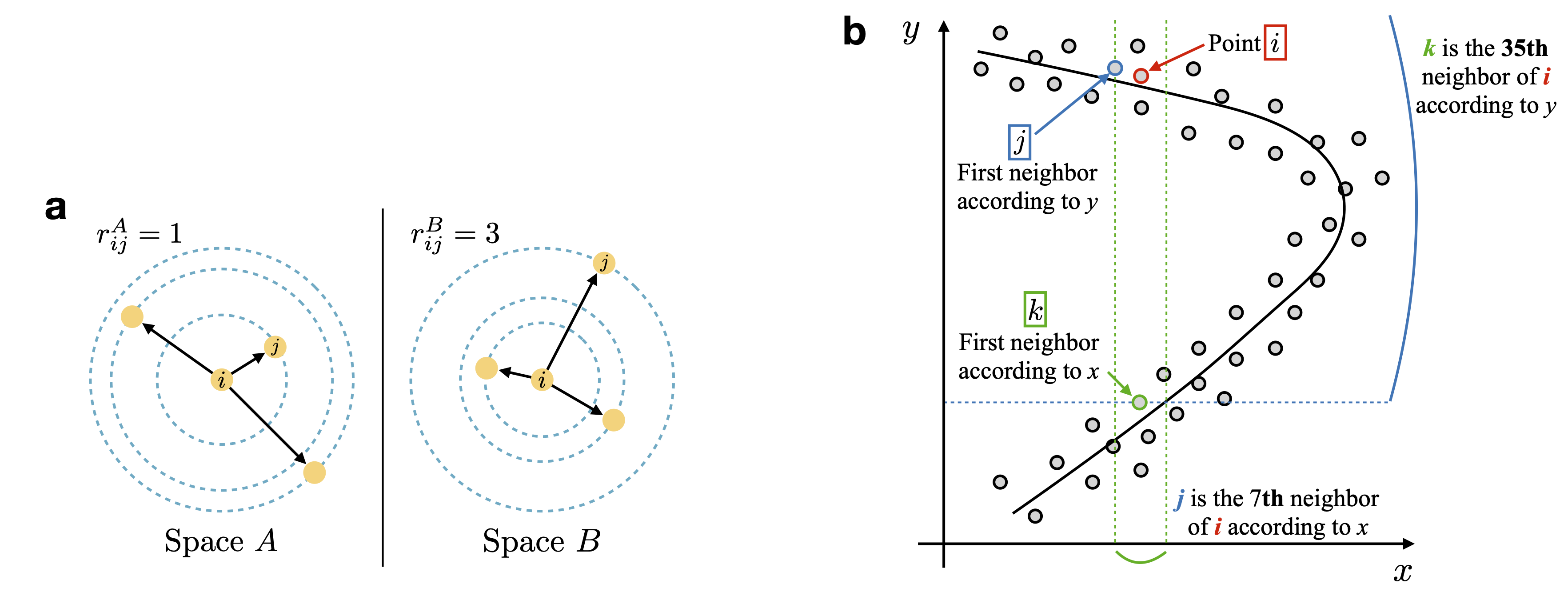}
	\caption{
	\textbf{Distance ranks can be used to measure the relative informations contained in two distance measures.}
	a): Illustration of the distance rank of two points in different feature spaces $A$ and $B$.
	The rank $r_{ij}$ of point $j$ relative to $i$ is equal to 1 in space $A$, meaning that $j$ is the first neighbor of $i$.
	This is not the case in space $B$, where point $j$ is the third neighbor of point $i$.
	b): Illustration of how ranks can be used to verify that space $x$ is less informative than space  $y$.
	The figure shows how a distance bound in $y$  automatically implies a distance bound in  $x$. The opposite is not necessarily true: the first neighbor of a point in the $x$ space can be very far in $y$.
}	
	\label{fig:rank_illustration}
\end{figure*}

In some cases, explicit features are not available, as in the case of raw text documents or genome sequences.
What one can always define, even in these cases, are \emph{distances} between data points whose definition, however, can be rather arbitrary \cite{Kaya:2019fo,Kulis:2013kz}.
How can one select, among an enormous amount of possible choices, the most appropriate distance measure for a given task?
Finding the correct distance is of course as difficult as performing feature selection or dimensionality reduction. 
In fact, these tasks can be considered equivalent if explicit features are available, since in this case a particular choice of features naturally gives rise to a different distance function computed through the Euclidean norm.

In this work, we approach feature/distance learning through a novel statistical and information theoretic concept.
We pose the question: given two distance measures $A$ and $B$, can we identify whether one is more \emph{informative} than the other?
If distance $A$ is more informative than distance $B$, even partial information on the distance $A$ can be predictive about $B$, while the reverse will not necessarily be true.
If this happens, and if the two distances have the same complexity e.g, they are built using the same number of features, $A$ should be generally preferred with respect to $B$ in any learning model. 

We introduce the concept of ``information imbalance'', a measure able to quantify the relative information content of one distance measure with respect to another.
We show how this tool can be used for feature learning in different branches of science.
For example, by optimizing the information content of a distance measure we are able to select from a set of more than 300 material descriptors, a subset of around 10 which is sufficient to define the state of a material system, and predict its energy. 
Moreover, we use the information imbalance to verify that the national policy measures implemented to contain the Covid-19 epidemic are informative about the future state of the epidemic.
In this case, we also show that the method can be used to detect \emph{causality} relationship between variables.

\section{The information imbalance}

Inspired by the widespread idea of using local neighborhoods to perform dimensional reduction \cite{Hastie2001_Elements_of_stat_learning} and classification \cite{Gashler2008_Sculpting} we quantify the relative quality of two distance measures by analyzing the \emph{ranks} of the first neighbors of each point.
For each pair of points $i$ and $j$, the rank $r_{ij}$ of point $j$ relative to point $i$, is obtained by sorting the pairwise distances between $i$ and rest of the points from smallest to largest.
For example, $r_{ij}^{A}=1$ if point $j$ is the first neighbor of point $i$ according to the distance $d_A$.
The rank of two points will be, in general, different when computed using a different distance measures $B$, as illustrated in Figure~\ref{fig:rank_illustration}a.

The key idea of our approach is that distance ranks can be used to identify whether one metric is more informative than the other.
Take the example given in Figure \ref{fig:rank_illustration}b, depicting a schematic representation of a noisy curved dataset.
In this dataset the distance along the $y$-axis is clearly more informative than the one along the $x$-axis since one could construct a function able to predict $x$ from the knowledge of $y$, but not the opposite.
This asymmetry is well captured by the ranks between points.
Take for example point $i$ (red circle in the figure).
Its first neighbour according to the $y$-distance is $j$ (blue circle), while according to the $x$-distance (green lines) $j$ is the 7th neighbour of $i$ .
Conversely, the nearest neighbour of $i$ according to the $x$-distance is $k$ (green circle), which is the 35th neighbour of $i$ according to the $y$-distance (blue lines).
We hence find that $r_{ij}^x \ll r_{ik}^{y}$ i.e., the rank in space $x$ of the first neighbor measured in space $y$ is much smaller than the rank in space $y$ of first neighbor measured in space $x$.

To give a more quantitative example, let's consider a dataset of points harvested from a 3-dimensional Gaussian whose standard deviation along the $z$ direction is a tenth of those along $x$ and $y$.
In this case, one can define a Euclidean distance between data points either using all the three features, $d_{xyz}^2=(x_i-x_j)^2+(y_i-y_j)^2+(z_i-z_j)^2$, or using a subset of these features ( $d_{xy}$, $d_{yz}$ and so on).

Intuitively, $d_{xyz}$ and $d_{xy}$ are almost equivalent since the standard deviation along $z$ is small, while there are information imbalances, say, between $d_{x}$ and $d_{xy}$, which would allow saying that $d_{xy}$ is more informative than $d_{x}$.
\begin{figure*}
    \centering
	\includegraphics[width=0.75\textwidth]{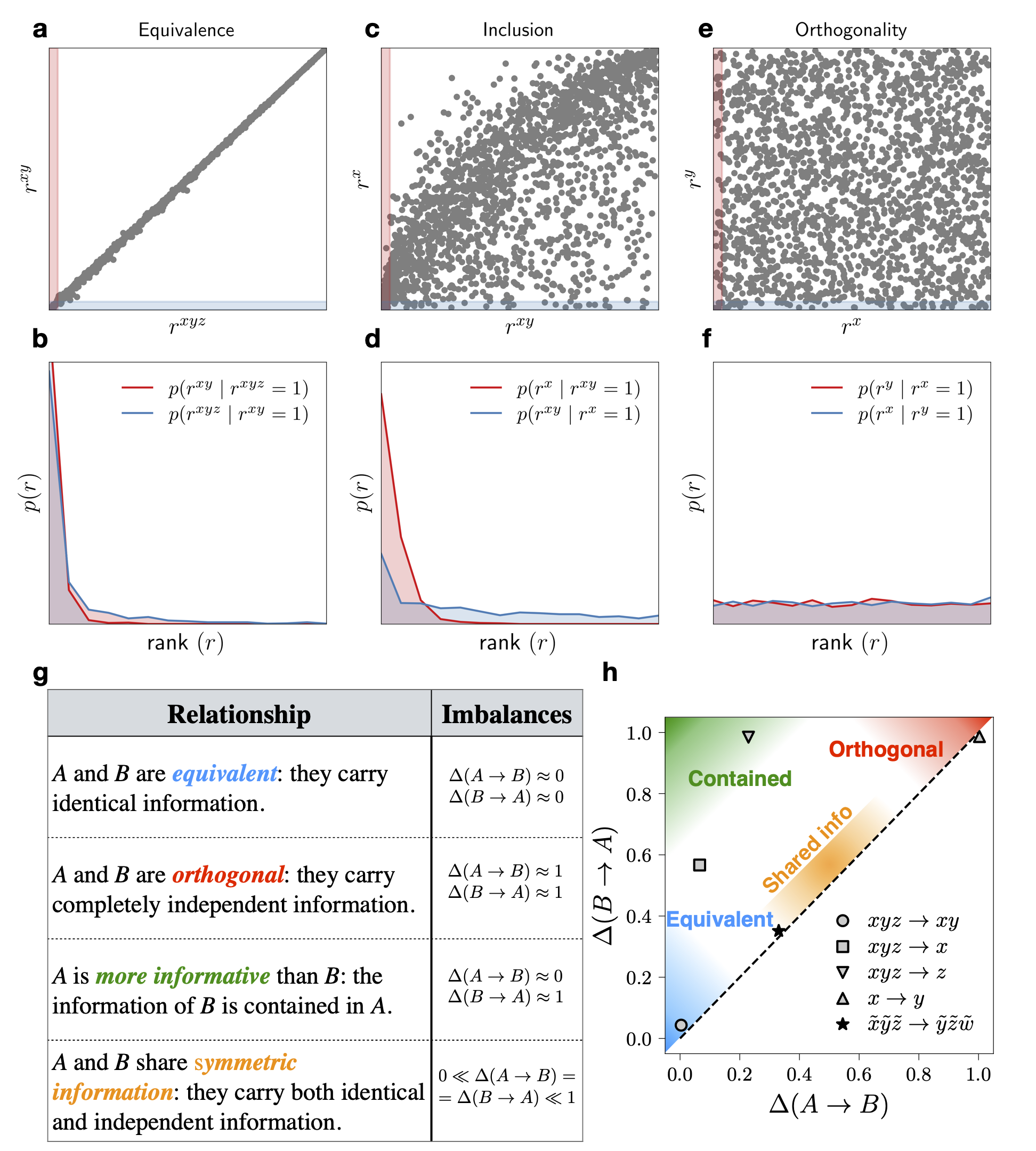}
	\caption{
	\textbf{Classifying and visualizing the relationship between any two distance measures using the information imbalance.}
a), c), e): Scatter plots of the ranks between ordered pairs of points for different distance measures evaluated on a 3D Gaussian dataset with a small variance along $z$.
The red and blue bands indicate, respectively, the points for which the rank on the $x$ and $y$ axis is equal to one (proportions were exaggerated for clarity).
The histogram of the points inside each band is plotted in the bottom plots using the same color.
b), d), f): Probability of that two nearest neighbor points ($r = 1$) for one distance have rank $r$ for the other distance. 
The three columns represent different pairs of representations.
g): The four different types of relationships that can characterize the relative information content of two spaces $A$ and $B$.
h): Information imbalance plane for the discussed 3D Gaussian dataset with small $z$-axis variance (gray markers), and for a 4D isotropic Gaussian dataset (black star).
The different colors (light blue, red, green and orange) roughly mark the regions corresponding to the four types of relationships listed in panel g.}
\label{fig:illustration_on_3dgauss}
\end{figure*}
In the first row of Figure \ref{fig:illustration_on_3dgauss}, we plot the ranks computed using one distance against the ranks computed using a second distance (for example the ranks in $d_{xy}$ as a function of those in $d_{xyz}$ for panel a).
In the second row of the figure we show the probability distribution $p(r^{A} \mid r^{B} =1)$ of the ranks $r_{ij}^{A}$ in space $A$ restricted to those pairs $(i,j)$ for which $r_{ij}^{B}=1$, namely to the  nearest neighbors according to distance $B$. 
In panels a and b, we compare the most informative distance containing all three coordinates to the one containing only the $x$ and $y$ coordinates.
Given the small variance along the $z$ direction, these two distance measures are practically equivalent, and this results in rank distributions strongly peaked around one.
In panels c and d, we compare the two metrics $d_{xy}$ and $d_x$.
In this case, the former is clearly more informative than the latter, and we find that the distribution of ranks when passing from $d_{xy}$ to $d_{x}$ is more peaked around small values than when going in the opposite direction. 
Finally, for two metrics built using independent coordinates ($x$ and $y$, in panels c and f) the rank distributions are completely uniform.

We hence propose to assess the relationship between any two distance measures $d_A$ and $d_B$ by using the properties of the conditional rank distribution $p(r^B \mid r^A = 1)$.
The closer this distribution is to a delta function peaked at one, the more information about space $B$ is contained within space $A$. 

This intuition can be made more rigorous through the statistical theory of copula variables.
We can define a copula variable $c_A$ as the cumulative distribution $c_A = \int_0^{d_A} p_A(w \mid x) dw$, where $p_A(w \mid x)$ is the of probability of sampling a data point within distance $w$ from $x$ in the $A$ space.
The value of $c_A$ can be estimated from a finite dataset by counting the fraction of points that fall within distance $d_A$ of point $x$, $c_A \approx r_A/N$. 
Copula variables and distance ranks 
can be considered continuous-discrete analogues of each other.
As a consequence, the distributions $p(r^{B} \mid r^{A} =1)$ shown in Figure~\ref{fig:illustration_on_3dgauss} are nothing else but estimates of the copula distributions $p(c_{B} \mid c_{A})$ with $c_A$ conditioned to be very small.
This is important, since Sklar's theorem guarantees that the copula distribution $p(c_A, c_B)$ contains the entire correlation structure of the metric spaces $A$ and $B$, independently of any details of the marginal distributions $p(d_A \mid x)$ and $p(d_B \mid x)$ \cite{Nelsen2006_Introduction_to_Copulas,Vincente_An_information_theoretic_approach,Panzeri_Information_estimation}.

Using the copula variables, we define the ``information imbalance'' from space $A$ to space $B$ as
\begin{equation}
	\Delta(A\rightarrow B) = 2 \lim_{\epsilon \rightarrow 0} \, \langle c_B \mid c_A = \epsilon \rangle ,
	\label{eq:imbalance_definition_copulas}
\end{equation}
where we used the conditional expectation 
$\langle c_B \mid c_A = \epsilon \rangle = \int c_B \, p(c_B \mid c_A = \epsilon)  dc_B$ 
to characterize the deviation of $p(c_B \mid c_A = \epsilon)$ from a delta function.
In the limit cases where the two spaces are equivalent or completely independent we have that $\langle c_B \mid c_A = \epsilon \rangle = \epsilon$ and $\langle c_B \mid c_A = \epsilon \rangle = 1/2$ respectively, so that the definition provided in Eq.~(\ref{eq:imbalance_definition_copulas}) statistically confines $\Delta$ in the range $(0,1)$.
The information imbalance defined in Eq.~(\ref{eq:imbalance_definition_copulas}) is estimated on a dataset with $N$ data points as 
\begin{equation}
    \Delta(A\rightarrow B) \approx 2\langle r^B \mid r^A = 1 \rangle/N
\end{equation}
We remark that the conditional expectation used in Eq.~(\ref{eq:imbalance_definition_copulas}) is only  one  of  the  possible quantities that can be used to characterize the deviation of the conditional copula distribution from a delta function.
Another attractive option is the entropy of the distribution.
In the Supplementary Material (SM) (S1.C), we show how these two quantities are related and we demonstrate that the specific choice does not substantially affect the results.
In the SM (see S1.B and Figure S1), we also show how copula variables can be used to connect the information imbalance to the standard information theoretic concept of mutual information.

By measuring the information imbalances $\Delta(A \rightarrow B)$ and $\Delta(B\rightarrow A)$, we can identify four classes of relationships between the two spaces $A$ and $B$.
We can find whether $A$ and $B$ are equivalent or independent, whether they symmetrically share both independent and equivalent information, or whether one space contains the information of the other.
These relationships are presented in Figure \ref{fig:illustration_on_3dgauss}g.
These relationships can be identified visually by plotting the two imbalances $\Delta(A\rightarrow B)$ and $\Delta(B\rightarrow A)$ against each other in a graph as done in Figure \ref{fig:illustration_on_3dgauss}h.
We will refer to this kind of graphs as \emph{information imbalance planes}.
In Figure~\ref{fig:illustration_on_3dgauss}h we present the information imbalance plane of the 3-dimensional Gaussian dataset discussed so far, and used for Figure~\ref{fig:illustration_on_3dgauss}a-f.
Looking at this figure, one can immediately verify that the small variance along the $z$ axis makes the two spaces $xyz$ and $xy$ practically equivalent (gray circle).
Similarly, one can verify that space $x$ is correctly identified to be contained in $xyz$ (gray square) and that the two spaces $x$ and $y$ are classified as orthogonal (gray triangle).
The figure also includes a point corresponding to a different dataset sampled from a 4-dimensional isotropic Gaussian with dimensions $\tilde{x}$, $\tilde{y}$, $\tilde{z}$ and $\tilde{w}$.
This point (black star) shows that the spaces $\tilde{x}\tilde{y}\tilde{z}$ and $\tilde{y}\tilde{z}\tilde{q}$ are correctly identified as sharing symmetric information. 
Importantly, the information imbalance only depends on the local neighborhood of each point and, for this reason, it is naturally suited to analyze data manifolds which are arbitrarily nonlinear.
In the SM (see S2.A and Figure S2), we show that our approach is able to correctly identify the best feature for describing a spiral of points wrapping around one axis, and a sinusoidal function.
More numerical tests are available online at \cite{DADApy} along with the corresponding code \cite{DADApy_paper}.
In the examples discussed so far we have chosen the Euclidean metric as distance measure for any subset of coordinates considered. 
We will make the same choice throughout the rest of this work.

\begin{figure*}[t]
    \centering
\includegraphics[width=0.93\textwidth]{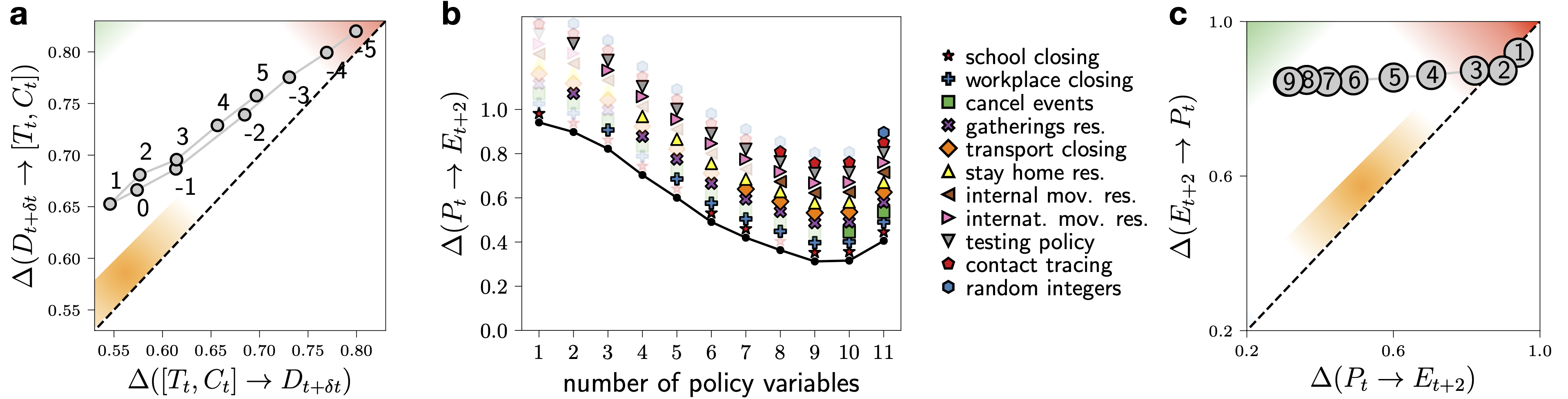}
    \caption{
    \textbf{The information imbalance allows for a straightforward verification of the impact of national policy measures on the Covid-19 epidemic.}
    a) Information imbalance between the space $[T_t, C_t]$ of the number of tests  and number of confirmed cases $C_t$ in a given week $t$, and the number of deaths $D_{t+\delta t}$ occurred in week $t + \delta t$. The time shift $\delta t$ ranges from $-5$ (5 weeks before) to $+5$ (5 weeks after).
    b): Minimum information imbalances from growing sets of policy variables $P_t$ to the state of the epidemic after two weeks $E_{t+2}$.
    c): The corresponding information imbalance plane with the number of policy variables going from 1 to 10 reported in the gray circles. 
    Point 10 is not visible as it lies below point 9.
    The figure shows that the policy measures space $P_t$ can predict the state of the epidemic $E_{t+2}$, while $E_{t+2}$ cannot predict $P_t$.
    }
    \label{fig:covid_fig}
\end{figure*}

\section{Influence of national policy measures on the \mbox{Covid-19} epidemic}

We now use the information imbalance to analyze 
the ``Covid-19 Data Hub'', a dataset which  provides comprehensive and up to date information on the Covid-19 epidemic~\cite{Guidotti:2020gz}, including epidemiological indicators such as the number of confirmed infections and the number of Covid-19 related deaths for nations where this is available, as well as the policy indicators that quantify the severity of the governmental measures such as school and workplace closing, restrictions on gatherings and movements of people, testing and contact tracing~\cite{Hale:wd}.
More details on the dataset are available in the SM (S2.B.1).

We first illustrate how the information imbalance can be used to recover the arrow of time from time series data.
In Figure \ref{fig:covid_fig}a we show the information imbalance between the space $[T_t,C_t]$, containing the number of tests $T_t$ and the number of confirmed cases $C_t$ in a given week~$t$, and the space of the number of deaths occurring in week $t+\delta t$ ($D_{t+\delta t}$). 
The imbalance is shown as a function of  $\delta t$.
All the points lay above the diagonal, indicating that, in the language of Figure~\ref{fig:illustration_on_3dgauss}h, the number of deaths is marginally \emph{contained} in the two variables $[T_t, C_t]$ if $\delta_t$ is small; and the optimal information imbalance occurs at $\delta t = 1$.
Importantly, for each pair of opposite time lags ($\delta t, - \delta t$) we find that the two variables $[T_t, C_t]$ always contain more information on future deaths than on past deaths. 
In this scenario this result represents an obvious verification of the known arrow of time of the dataset, but it suggests that further dedicated investigations could bring to the development of accurate tests to detect nontrivial causality relationships~\cite{inferring_causality}.

We now analyze the relationship between  the space of policy measures $P_t$ at week $t$, and the state of the epidemic $E_{t + \delta t}$, with $\delta t= 2$ (namely after  two weeks). In SM (S2.B.2, Figure S3) we show that the analysis with time lags of one or three weeks bring to similar results.
While we consider several different choices for the policy space, the epidemic state is defined by the number of weekly deaths $D_{t}$ and the ratio $R_{t} = C_{t}/T_{t}$ of confirmed cases $C_{t}$ over total number of tests performed $T_{t}$ per week.
We estimate the information imbalance $\Delta(P_t \rightarrow E_{t + \delta t})$ between the spaces defined by all the possible combination  of policy measures  $P_t$  and the space of epidemiological variables $E_{t + \delta t}$. A low value of $\Delta(P_t \rightarrow E_{t+ \delta t})$ means that $P_t$ can predict $E_{t + \delta t}$.
In Figure \ref{fig:covid_fig}b we present the minimum information imbalance $\Delta(P_t \rightarrow E_{t + \delta t})$ achievable with any set of $d$ policy measures.

For $d \le 2$, $\Delta(P_t \rightarrow E_{t + \delta t})$ is close to one, indicating that no single or couple of policy measure is  significantly predictive about the state of the epidemic, consistently with \cite{Haug_Ranking_the_effectiveness}.
When three or more policy measures are considered, the information imbalance decreases rapidly reaching a value of about $0.28$ when almost all policy measures are considered.
This sharp decrease and the low value of the information imbalance clearly indicate that policy measures \emph{do contain} information on the future state of the epidemic, and the more policy measures are considered, the more the future state of the epidemic can be considered as \emph{contained} in the space of the policies. 
As a sanity check, a dummy policy variable was introduced for this test (blue hexagon).
This variable is never selected by the algorithm, and its addition deteriorates the information content of the policy space.
The described analysis verifies that
policy interventions have been effective in containing the spreading of the Covid-19 epidemic, a result which has been shown in a number of studies~\cite{Brauner_Inferring_the_effectiveness,Haug_Ranking_the_effectiveness,Hsiang_The_effect_of_large_scale,Flaxman_Estimating_the_effects}.
In accordance with these studies, we also find that multiple measures are necessary to effectively contain the epidemic, with no single policy being sufficient on its own \cite{Soltesz_Matters_arising}, and that the impact of policy measures increases monotonically with the number of measures put in place.
We find that a small yet effective set of policy measures has been the combination of testing, stay home restrictions and restrictions on international movement and gatherings.
While our results are computed as averages over all nations considered, further analysis carried out in the SM (S2.B.3, Figure S5) on disjointed subsets of nations give results which are consistent with our main findings.

We finally note that the information imbalance $\Delta(E_{t+2} \rightarrow P_t)$ (shown in Figure \ref{fig:covid_fig}c) remains considerably high for any number of policy variables.
This indicates that the state of the epidemic is not informative about past policy measures. 
Surprisingly, the state of the epidemic is not informative even on \emph{future} policy measures (see S2.B.2 and Figure S4 of the SM), a result which seems to indicate that that different nations have reacted to the epidemic through widely different strategies.

The information imbalance can also be used to optimally choose the relative scale of heterogeneous variables.
For instance, in the SM (S2.B.4, Figure S6), we use the information imbalance to select the relative scale of heterogeneous epidemiological variables such as the total number of tests and the ratio of confirmed cases over total number of tests. 
This is important in real-world applications, where features are often characterized by different units of measure and different scales of variations.

\begin{figure*}
\includegraphics[width=1.0\textwidth]{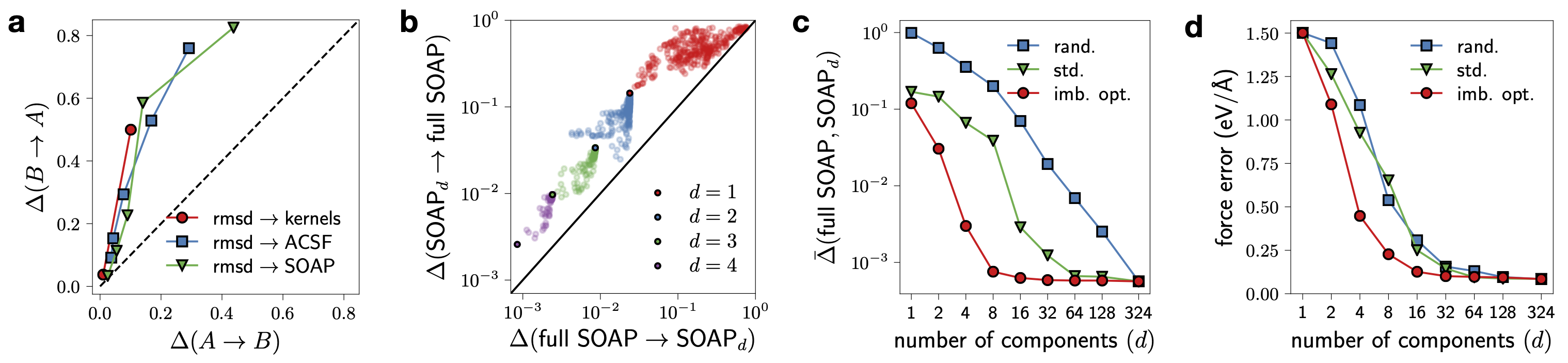}
	\caption{
	\textbf{The information imbalance can be  used to perform an effective information-driven feature selection in materials systems.}
	a): Information imbalances between ground truth ``rmsd'' distance metric and standard atomistic descriptors.
	b): Information imbalances between a full description and the most informative $d$-plet of components ($d=1,\dots,4$). %
	c): Convergence of the ``symmetric'' information imbalance with the number of components for three different compression strategies. 
	The symmetric information imbalance is defined as $\bar{\Delta} (A, B) = [\Delta(A \rightarrow B) + \Delta(B \rightarrow A)] /\sqrt{2}$; more details can be found in the SM (S2.C.3).
	d): Force error on a validation set of a machine learning potential energy model built on the compressed descriptors.
	}
	\label{fig:test_materials}
\end{figure*}

\section{Selection and compression of descriptors in materials physics}

We now show that the information imbalance criterion can be used to assess the information content of commonly used numerical descriptors of the geometric arrangement of atoms in materials and molecules, as well as to compress the dimension (number of features) of a given descriptor with minimal loss of information.
Such atomistic descriptors are needed for applying any statistical learning algorithm to problems in physics and chemistry \cite{Physics:vk,Schutt:2020ww,Carleo2019_RevModPhys,Schmidt:2019iz,Butler:2018fla}, and the problem of choosing optimally informative atomic descriptors has recently attracted attention \cite{Ceriotti_inf_loss}.
We first consider a database consisting of an atomic trajectory of amorphous silicon generated from a molecular dynamics simulation at 500$\rm{K}$
(see S2.C.1 of the SM for details).
At each time step of this trajectory we select a single local environment by including all the neighboring atoms within the cutoff radius of $4.5\text{\AA}$ from a given central atom.
In this simple system, which does not undergo any significant atomic rearrangement, one can define a fully informative distance measure as the minimum over all rigid rotations of the root mean square deviation (rmsd) of two local environments (details in S2.C.2 of the SM).

In Figure \ref{fig:test_materials}a, this ground truth distance measure is compared with some of the descriptors most commonly used for materials modeling: the ``Atom-centered Symmetry Functions'' (ACSF) \cite{Behler:2007fe,Behler:2011ita}, the ``Smooth Overlap of Atomic Positions'' (SOAP) \cite{Bartok:2013cs,Caro:2019eta} and the 2 and 3-body kernels \cite{Glielmo:2018bm,Zeni:2019hg}.
Unsurprisingly, all descriptors are contained in the ground truth distance measure.
For ACSF and SOAP representations, one can increase the resolution by increasing the size of the descriptor in a systematic way, and we found that doing this allows both representations to converge to the ground truth.

A materials descriptor typically involves a few hundred components. 
Following a procedure similar to the one used in the last section to select policy measures, we use the information imbalance to efficiently compress these high-dimensional vectors with minimal loss of information (more details are given in S2.C.3 of the SM).
We perform this compression for a database consisting of complex geometric arrangements of carbon atoms \cite{Deringer:2017ea}. 
%
As illustrated in Figure \ref{fig:test_materials}b and c, the selection leads to a rapid decrease of the information imbalance, and converge much more quickly than other strategies such as random selection (blue squares) and standard sequential selection (green triangles).
Figure \ref{fig:test_materials}d depicts the test error of a potential energy model constructed using a state-of-the-art Gaussian process regression model \cite{Bartok:2010fj} (see S2.C.5 of the SM) on the compressed descriptors, as a function of the size of the descriptors and for the different compression strategies considered.
Remarkably, the graph shows that a very accurate model can be obtained using only 16 out of the 324 original components of the descriptor considered \cite{Caro:2019eta}.
Figures \ref{fig:test_materials}c and \ref{fig:test_materials}d show that when the information imbalance has converged, the validation error does not diminish further. 
This suggests that one can select the optimal descriptor dimension as the one for which no improvement in the information imbalance is observed. In the SM (S2.C.6, Figure S8) we show how a similar criterion can be also used to select the hyper-parameters of materials descriptors, and we demonstrate how the order of the components selected by our procedure can be understood considering the fundamental structure of the descriptor.
In the SM (S2.C.7, Figure S7) we show that, for this prediction task, the feature selection scheme based on the information imbalance is marginally more efficient than other well known compression schemes for materials descriptors.

\section{Conclusions}

In this work we introduce the information imbalance, a new method to assess the relative information content between two distance measures.
The key property which makes the information imbalance useful is its asymmetry:
it is different when computed using a distance $A$ as a reference and a distance $B$ as a target, and when the two distances are swapped.
This allows distinguishing three classes of similarity between two distance measures: a full equivalence, a partial but symmetric equivalence, and an asymmetric equivalence, in which one of the two distances is observed to contain the information of the other.

The potential applications of the information imbalance criterion are multifaceted.
The most important one is probably the long-standing and crucial problem of feature selection~\cite{vanderMaaten:2008tm, mcinnes2018umap, Bengio:2013bu}.
Low-dimensional models typically allow for more robust predictions in supervised learning tasks \cite{Lopes2017_Facial_expression_few_data,Yang2018_Feature_selection_new_perspective}. Moreover, they are generally easier to interpret and can be used for direct data visualization if sufficiently low dimensional.
We design feature selection algorithm by selecting the subset of features which minimizes the information imbalance with respect to a target property, or to the original feature space.

As we have showcased, such algorithms can be ``exact'' if the distances to be compared are relatively few (as done for the Covid-19 database) or approximate, if one has to compare a very large number of distances (as done for the atomistic database).
Such algorithms work well even when in the presence of strong nonlinearities and correlations within the feature space.
This is exemplified by the analysis of the Covid-19 dataset, where 4 policy measures which appear similarly irrelevant when taken singularly, were instead identified as maximally informative when taken together with regards to the future state of the epidemic.\\
Other applications include dimensionality reduction schemes that directly use the information imbalance as an objective function.
Admittedly such function will in general be non differentiable and highly non-linear, but efficient optimization algorithms could still be developed by exploiting recent results on the computation of approximate derivatives for sorting and ranking operations \cite{pmlr-v119-blondel20a}.\\
Another potentially fruitful line of research would be exploiting the information imbalance to optimize the performance of deep neural networks.
For example, in SM (S2.C.8, Figure S9), we show that one can reduce the size of the input layer of a neural network that predicts the energy of a material, yielding more computationally efficient and robust predictions.
However, one can imagine to go much further, and compare distance measures built using the representations in different hidden layers, or in different architectures.
This could allow for designing maximally informative and maximally compact neural network architectures.
We finally envision potential applications of the proposed method in the study of causal relationships: we have seen that in the Covid-19 database the use of information imbalance makes it possible to distinguish the future from the past, as the former contains information about the latter, but not vice-versa.  
We believe that this empirical observation can be made robust by dedicated theoretical investigations, and used in practical applications in other branches of science.

\section{Acknowledgments}
AG, CZ, and AL gratefully acknowledge support from the European Union’s Horizon 2020 research and innovation program (Grant No. 824143, MaX 'Materials design at the eXascale' Centre of Excellence).
The authors would like to thank M. Carli, D. Doimo and I. Macocco (SISSA) for discussions, M. Caro (Aalto University) for precious help in using the TurboGap code and D. Frenkel (University of Cambridge) and N. Bernstein (U.S. Naval Research Laboratory) for useful feedback on the manuscript.

\section{Supplementary Material}
Supplementary material is available at PNAS Nexus online.

\section{Funding}
This work is supported in part by funds from the European Union’s Horizon 2020 research and innovation program (Grant No. 824143, MaX 'Materials design at the eXascale' Centre of Excellence).

\section{Author contributions statement}
A.L., A.G., G.C. and B.C. designed research; A.G performed research; A.L. and C.Z. contributed to perform research; A.G., C.Z., B.C., G.C. and A.L. wrote the paper.


\section{Data availability}
Details on the datasets used are available in the supplementary material.

\bibliographystyle{unsrtnat}
\bibliography{reference}

\end{document}